\documentclass{article}
\usepackage{ijcai19}

\usepackage{amsmath,amsfonts,bm}

\def\eqref#1{equation~\ref{#1}}

\def\1{\bm{1}}

\def\vg{{\bm{g}}}
\def\vh{{\bm{h}}}

\def\vl{{\bm{l}}}

\def\vx{{\bm{x}}}

\DeclareMathAlphabet{\mathsfit}{\encodingdefault}{\sfdefault}{m}{sl}
\SetMathAlphabet{\mathsfit}{bold}{\encodingdefault}{\sfdefault}{bx}{n}

\def\gD{{\mathcal{D}}}

\def\gG{{\mathcal{G}}}

\newcommand{\E}{\mathbb{E}}

\usepackage{times}
\usepackage{soul}
\usepackage{url}
\usepackage[hidelinks]{hyperref}
\usepackage[utf8]{inputenc}
\usepackage[small]{caption}
\usepackage{graphicx}
\usepackage{amsmath}
\usepackage{booktabs}
\usepackage{algorithm}
\usepackage{algorithmic}
\title{Feature Prioritization and Regularization Improve Standard Accuracy and Adversarial Robustness}

\author{
Chihuang Liu\footnote{Contact Author}\And
Joseph JaJa\\
\affiliations
Institute for Advanced Computer Studies and Department of Electrical and Computer Engineering\\
University of Maryland, College Park, MD 20742, USA\\
\emails
\{chliu, josephj\}@umd.edu
}

\begin{document}

\maketitle

\begin{abstract}
Adversarial training has been successfully applied to build robust models at a certain cost. While the robustness of a model increases, the standard classification accuracy declines. This phenomenon is suggested to be an inherent trade-off. We propose a model that employs feature prioritization by a nonlinear attention module and $L_2$ feature regularization to improve the adversarial robustness and the standard accuracy relative to adversarial training. The attention module encourages the model to rely heavily on robust features by assigning larger weights to them while suppressing non-robust features. The regularizer encourages the model to extract similar features for the natural and adversarial images, effectively ignoring the added perturbation. In addition to evaluating the robustness of our model, we provide justification for the attention module and propose a novel experimental strategy that quantitatively demonstrates that our model is almost ideally aligned with salient data characteristics. Additional experimental results illustrate the power of our model relative to the state of the art methods.
\end{abstract}

\section{Introduction}
\label{section_introduction}

Deep learning models have demonstrated impressive performance in a wide variety of applications~\cite{goodfellow2016deep,simonyan2014very}. However, recent works have shown that these models are susceptible to adversarial attacks: imperceptible but carefully chosen perturbation added to the input can cause the model to make highly confident but incorrect predictions ~\cite{szegedy2013intriguing,goodfellow2015explaining,kurakin2016adversarial}.

Exploring the adversarial robustness of neural networks has recently gained significant attention and there is a rapidly growing body of work related to this topic~\cite{kurakin2016adversarial,tramer2017ensemble,fawzi2018analysis,athalye2017synthesizing,carlini2017towards,kolter2017provable,wong2018provable,madry2017towards}. A wide variety of methods are proposed to defend a model against adversarial attacks~\cite{prakash2018deflecting,liao2018defense,song2017pixeldefend,samangouei2018defense}. Despite these 
advances, many techniques are subsequently shown to be ineffective~\cite{athalye2018obfuscated,athalye2018robustness}, and adversarial training which uses adversarial samples in addition to clean images during the training process has been shown to be able to build relatively robust neural networks~\cite{madry2017towards,athalye2018obfuscated,dvijotham2018training}. With strong adversaries such as the Projected Gradient Descent (PGD)~\cite{madry2017towards} or the Iterative Fast Gradient Sign Method (I-FGSM)~\cite{kurakin2016adversarial} adversarially trained models are able to achieve state-of-the-art performance against a wide range of attacks.

Recent advances in the understanding of adversarial training provide insights of its effectiveness. It is shown that standard and robust models depend on very different sets of features~\cite{tsipras2018there,tanay2018built}. While standard models utilize features including non-robust ones that are weakly correlated with class labels and easily manipulated by small input perturbations, robust models only use robust features that are highly correlated with class labels and invariant to those perturbations. Although adversarial training learns robust features, there are no explicit design components to encourage a model to depend solely on robust features. Therefore, to further improve the robustness of a model, we propose feature regularization and prioritization schemes.

We first propose to use an attention mechanism that introduces to the model the flexibility of prioritizing features and bias a model towards robust features. We call the learned features at the final layer of a network the global features, and the ones at lower level layers the local features. In our attention module, we use the global features as a way to assign weights to the local features by a non-linear compatibility function. Since global features are directly used to produce class label prediction, we are effectively assigning weights to local features depending on their correlation with the labels. Robust features have higher correlation and therefore will be assigned larger weights which in turn contribute to the model's robustness.

Next, we propose to use feature regularization to learn robust features that are invariant to input perturbations. We add an $L_2$ regularization term that penalizes the distance between the learned features of a clean sample $x$ and that of its perturbed adversarial counterpart $x'$ to the training objective. By optimizing this regularizer, we are pushing the model to extract very similar features from the original image and the adversarial image, and thus only features that are invariant to the perturbations are learned and the added noise is effectively ignored. From another point of view, a model with small $L_2$ feature distance maps the two nearby points in the image space to nearby points in the learned high dimensional manifold, which is a desirable behavior.

In this paper, we propose an approach that enhances adversarial training with feature prioritization and regularization to improve the robustness of a model. We use extensive experiments to demonstrate that the attention module focuses on the area of an image which contains the actual object and helps the classifier to only rely on features extracted from those areas. The background clutter and irrelevant features which could be misleading are effectively suppressed. The feature regularization further encourages the model to extract robust features that are not manipulated by the adversarial perturbations. The resulting model has a highly interpretable gradient map that aligns very well with salient data characteristics.

The main contributions of this paper include:
\begin{itemize}
    \item A method based on feature prioritization and regularization, which significantly outperforms adversarial training. Our model is evaluated on the MNIST, CIFAR-10, and CIFAR-100 datasets, and demonstrates superior performance relative to both standard classification accuracy and adversarial robustness.
    \item We provide empirical justification for attention and show that it helps the model to rely on robust features by assigning larger weights to them. Through qualitative inspection, we show that the attention maps generated by our non-linear attention estimator focus sharply on the regions of interest while suppressing irrelevant background clutter.
    \item In addition to qualitative evaluation of the gradient maps, we propose a novel experimental strategy that quantitatively demonstrates better alignment of the gradient maps generated by our model with salient data characteristics.
\end{itemize}

\section{Related Work}
\label{section_related_work}
Due to the extensive amount of literature in this area, we only review some of the most related works in this section. For a more comprehensive survey, please refer to Akhtar and Mian~\shortcite{akhtar2018threat}. 

\textit{Adversarial training.} Kurakin et al.~\shortcite{kurakin2016adversarial} use adversarial training as a form of data augmentation where it injects adversarial examples during training. In every training mini batch, a mixture of clean images and adversarial images generated by one step Fast Gradient Sign Method (FGSM) are used to update the network's parameters. It was later improved by Na et al.~\shortcite{na2017cascade} by adding adversarial examples generated by iterative methods. Madry et al.~\shortcite{madry2017towards} proposed to replace all clean images with adversarial images which is a direct result of optimizing a saddle point (min-max) formulation. They suggest that PGD is a universal first-order adversary which is then used in their adversary generating process.

\textit{$L_2$ regularization}. A similar idea with feature regularization is proposed in  Kannan et al.~\shortcite{kannan2018adversarial} which they call adversarial logit pairing (ALP), to  prevent a model from being over-confident when making predictions. Compared with ALP, feature regularization is more intuitive as it motivates a model to learn very robust features that are invariant to input perturbations, which leads to a robust model. In addition, we propose to also use attention module to further encourage the model to favor robust features which will improve the robustness.

\textit{Attention Models}. Attention in CNN is most commonly deployed for query-based tasks~\cite{seo2016progressive,jetley2018learn}. Jetley et al.~\shortcite{jetley2018learn} presented a method to use a learned representation of the global image as a query to leverage multiple attention maps at different scales, which allows the expression of a complementary focus on different parts of the image. However, the application of attention to the adversarial robustness aspects has not been seriously explored. To the best of our knowledge, we are the first to employ an attention mechanism in training a robust deep neural network. In our application, we use a ReLU activated neural network instead of the linear-based method as the attention estimator. It allows highly non-linear compatibility between the learned global features and the lower-level local features. 

\section{Approach}
\label{section_approach}
\begin{figure*}[t]
\begin{center}
\includegraphics[width=0.8\textwidth]{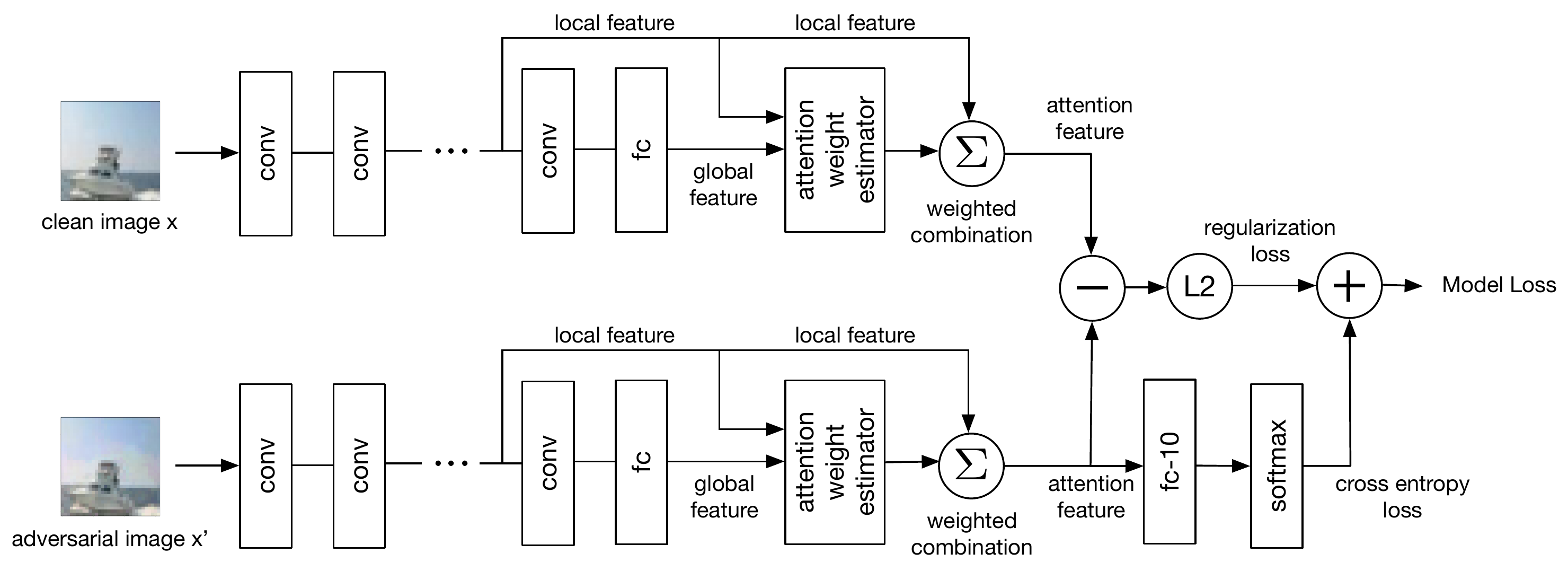}
\end{center}
\caption{Overview of the proposed model. The top and bottom networks are the same copy that share all network parameters. Both the clean and adversarial images are forwarded through the network to produce the corresponding attention features. The $L_2$ regularization loss is defined as the Euclidean distance between the two sets of attention features. The final model loss is a combination of the $L_2$ regularization loss and the cross-entropy loss for only the adversarial input. \label{figure_model}}
\end{figure*}

We now present our model that combines the attention module and $L_2$ feature regularization, and show how it can be applied to enhance the adversarial training to improve the adversarial robustness of a model and its accuracy. Figure~\ref{figure_model} provides an overview of our method. We start by forwarding each of the clean and adversarial images and computing the attention weights by a non-linear estimator. Then the individual attention feature is defined to be the weighted combination of the corresponding local features. Next, we define an $L_2$ regularization loss to be the Euclidean distance between the two sets of learned attention features. The attention features of the adversarial image are then used to produce the logits, which is followed by softmax layer to produce the cross-entropy loss. The final loss function of our model is a combination of cross-entropy loss and the regularization loss.

\subsection{Adversarial Training}
We adopt the adversarial training described in Madry et al.~\shortcite{madry2017towards} as the basic training approach. It replaces natural training examples by PGD examples, which is suggested to represent a universal first-order adversary. So far PGD has been shown to represent the strongest attack method \cite{athalye2018obfuscated,athalye2018robustness}. A model that is trained with PGD adversaries is also robust against a wide range of other attacks and not yet outperformed by any other approach. The adversarial training has a saddle point formulation:

\begin{equation}
\label{equation_pgd_saddle}
\min_\theta \E_{(\vx, y)\sim \gD} [ \max_{\bm\delta \in S} L(\theta, \vx+\bm\delta, y)]
\end{equation}
where $\gD$ is the distribution of data $\vx$ and class labels $y$, $L$ is the cross-entropy loss function for a model with parameters $\theta$, $\bm\delta$ is the additive adversarial perturbation with bound $S$. In this paper we consider $l_\infty$ bound as in Madry et al.~\shortcite{madry2017towards}. Our adversarial samples $\vx'=\vx+\bm\delta$ are created by PGD:
\begin{equation}
    \vx^{t+1}=\Pi_{\vx+S}\left(\vx^t+\alpha\,\text{sgn}(\nabla_\vx L(\theta, \vx, y))\right)
\end{equation}

PGD adversaries are computed at each iteration as an approximated optimum of the inner maximization in equation~(\ref{equation_pgd_saddle}) and an update of the parameters $\theta$ is made according to the outer minimization formulation.

\subsection{Attention Model}
As we discussed in Section~\ref{section_introduction}, our goal of attention model is to favor robust features in making predictions. We propose a non-linear attention model that acts as a feature prioritizing scheme, which is able to put more weight on robust features and less weight on non-robust features to increase the robustness of a classifier.

Let $\vl_n^i$  be the learned feature vector at layer $i\in\{1, 2, ..., I\}$ of a neural network at spatial location $n\in\{1,2, ..., N\}$, and let $\vg$  be the feature vector of the layer just before the final fully connected layer which produces the class label prediction scores (logits). We use a one-hidden-layer ReLU network to generate compatibility scores between the global feature $\vg$ and local features $\vl_n^i$:
\begin{equation}
    c_n^i = f(\vl_n^i, \vg)
\end{equation}
where $f$ is the neural network and the concatenation of $\vg$ and $\vl_n^i$ is fed to the network to produce the compatibility scores $c_n^i$. We then normalize the scores with a softmax operation to get the attention weights:
\begin{equation}
    w_n^i = \frac{\exp{c_n^i}}{\sum_{m}\exp{c_m^i}}
\end{equation}
Next, we compute the weighted sum of local feature vectors which is the attention feature vector at layer $i$:
\begin{equation}
    \vh^i = \sum_n w_n^i \vl_n^i
\end{equation}
We use the outputs of the last residual block as the local feature for computing attention, and replace the global feature $\vg$ with the corresponding attention descriptor $\vh^i$ for final classification. 

By using a neural network instead of the linear alignment models as in Jetley et al.~\shortcite{jetley2018learn}, we are able to capture non-linear compatibility between the local and global features when producing the attention weights, which is beneficial considering the multiple non-linear function activated layers between the local and global features.

\subsection{Feature Regularization}
In addition to the attention mechanism, we also propose to use an $L_2$ regularization term to encourage the model to extract similar features for the clean image and the corresponding adversarial image. Denote by $\gG_\theta$ the deep neural network, $\vx$ and $\vx'$ the natural image and adversarial image. Denote by $\gG_\theta(\vx)$, $\gG_\theta(\vx')$ the learned features of the layer just before the final fully connected layer (in our case this is the attention weighted global descriptor) which produces the class label prediction scores. The $L_2$ regularizer has the following form:
\begin{equation}
    L_r(\vx, \vx') = \|\gG_\theta (\vx) - \gG_\theta (\vx')\|_2
\end{equation}
By minimizing the regularization function, the model effectively learns very similar features for the clean sample and the adversarial sample, which are robust features since they are invariant to the adversarial perturbation. From another perspective, the learned features of a neural network lie on a high dimensional manifold that is linearly separable for different classes because the classification layer is a linear classifier followed by a softmax function. With adversarial training alone, a model only tries to map $\vx$ and $\vx'$ to the same side of the decision boundary, while with the additional regularization, they are not only on the same side but also mapped to nearby points in the space. This mapping is a desired behavior considering that, in the original image space, they are very close points representing essentially the same image.

\subsection{Model Loss}
Equipped with the presented methods, the total loss of our model is:
\begin{equation}
   \text{Loss}=\E_{(\vx, y)\sim \gD} [  L(\theta, \vx', y) + \lambda \|\gG_\theta (\vx) - \gG_\theta (\vx')\|_2]
\end{equation}
where $\lambda$ is a hyperparameter that controls the relative weight of the $L_2$ regularization loss.

\section{Experiments and Results}
\label{section_experiments_and_results}
In this section, we evaluate our model on the MNIST, CIFAR-10 and CIFAR-100 datasets, and present empirical justification to attention module and some quantitative and qualitative results.
\subsection{Robustness on MNIST}
\label{subsection_results_on_mnist}
We use a CNN with two convolutional layers with 32 and 64 filters respectively, followed by two fully connected layers of size 1024 and 10. The network is trained with  40-step PGD adversary with a step size of 0.01 and $l_\infty$ bound of $\epsilon=0.3$. The settings are the same as in Madry et al.~\shortcite{madry2017towards}. Since MNIST is a very small scale dataset and the model is very robust with just adversarial training, we do not employ the attention mechanism, but only study the effectiveness of the proposed feature regularization method.
\begin{table}[h]
\begin{center}
\begin{tabular}{@{}l@{\hskip 0.07\textwidth}r@{\hskip 0.03\textwidth}r@{\hskip 0.03\textwidth}r@{}}
\toprule
\bf Method & \bf Natural & \bf White box & \bf Transfer \\ \midrule
Madry et al.  & 98.72\%    & 92.86\%        & 95.97\%       \\
AT-reg  & \bf 98.97\%    & \bf 95.95\%        & \bf 96.90\%      \\ \bottomrule
\end{tabular}
\caption{Performance comparison of the adversarial training and our proposed adversarial training with feature regularization (AT-reg) on MNIST against PGD 5 adversaries. Transfer attack accuracies are evaluated against adversaries generated from an independently trained copy of the same method with identical configurations.\label{table_mnist_results}}
\end{center}
\end{table}

The evaluation results are presented in Table~\ref{table_mnist_results}. Regarding the value of weight $\lambda$ of the $L_2$ regularizer, we find that roughly any $\lambda\in[0.001,0.1]$ works well. The reported results are obtained with $\lambda=0.1$. Table~\ref{table_mnist_results} shows that a model trained with the proposed feature regularization method is significantly more robust against PGD adversary than the baseline model with adversarial training alone. The improvement is more than 3\% for white box attack and 1\% for transfer attack.

\subsection{Robustness on CIFAR-10}
\label{subsection_results_on_cifar10}

We use the same wide residual network as Madry et al.~\shortcite{madry2017towards} with [16, 16, 32, 64] filters and its 3-times wider variant with [16, 48, 96, 192] filters respectively. For our attention model, we modify the ResNet by replacing the spatial global average pooling layer after the residual block 4 with a convolutional layer sandwiched between two max-pooling layers to obtain the global feature $\vg$. We use a one-hidden-layer ReLU neural network with 64 hidden units as the non-linear attention weight estimator. The model is trained with 5-step PGD adversary and a step size of 2 and $l_\infty$ bound of $\epsilon=8/255$. In order to isolate and analyze the effectiveness of attention module and feature regularization independently, we train three models with the following configurations: adversarial training with feature regularization (AT-reg), adversarial training with attention (AT-att), and adversarial training with both attention and feature regularization (AT-att-reg). We then evaluate the models using PGD and CW~\cite{carlini2017towards} adversaries optimized with various number of steps under white box and transfer attack settings. Transfer attack accuracies are evaluated against adversaries generated from an independently trained copy of the same method with identical configurations.

\begin{table}[t]
\begin{center}
\begin{tabular}{@{}l@{\hskip -0.02\textwidth}r@{\hskip 0.015\textwidth}r@{\hskip 0.015\textwidth}c@{\hskip 0.002\textwidth}r@{}}
\toprule
\bf Method & \bf Madry et al.  & \bf AT-reg & \bf AT-att & \bf AT-att-reg  \\ \midrule
  Natural & 80.79\%    & 79.52\% & \bf 82.43\%   & 81.20\%\\\midrule
White, PGD 5  & 49.89\%  & 52.35\%  & 51.22\%  & \bf 53.38\%\\
White, PGD 20 &39.72\% & 44.25\% & 41.40\%&\bf 45.20\% \\
White, PGD 100 &38.76\% &43.73\% & 40.85\%&\bf 44.60\%\\
White, PGD 200 &38.64\% &43.70\% & 40.72\%&\bf 44.54\%\\
White, CW 30 &40.27\% &42.96\% & 40.60\%&\bf 44.37\%\\
White, CW 100 &39.98\% &42.87\% & 40.31\%&\bf 44.26\% \\\midrule
Transfer, PGD 5& 60.13\%& 61.82\% & \bf 63.26\% & 62.32\%\\
Transfer, PGD 20 &56.60\% &56.40\% &\bf 58.26\%&57.52\%\\
Transfer, PGD 100 &56.44\% &56.28\% &\bf 58.08\%&57.53\%\\
Transfer, PGD 200 &56.49\% &56.25\% &\bf 58.04\% &57.53\%\\
Transfer, CW 30 &57.11\% &56.86\% &\bf 58.07\%&57.57\%\\
Transfer, CW 100 &57.10\% &56.79\%&\bf 58.13\%&57.41\%\\ \bottomrule
\end{tabular}
\caption{Performance comparison of the adversarial training, adversarial training with feature regularization (AT-reg), adversarial training with attention (AT-att), and adversarial training with both (AT-att-reg) on CIFAR-10 against PGD and CW adversaries.\label{table_cifar10_narrow_results}}
\end{center}
\end{table}

\begin{table}[t]
\begin{center}
\begin{tabular}{@{}l@{\hskip -0.02\textwidth}r@{\hskip 0.015\textwidth}r@{\hskip 0.015\textwidth}c@{\hskip 0.002\textwidth}r@{}}
\toprule
\bf Method & \bf Madry et al.  & \bf AT-reg & \bf AT-att & \bf AT-att-reg  \\ \midrule
  Natural & 85.41\%    & 84.65\% & \bf 86.48\%   & 85.98\%\\\midrule
White, PGD 5  & 49.15\%  & 52.21\%& 50.91\%  & \bf 53.23\%\\
White, PGD 20 &38.19\% & 41.00\% & 39.52\%&\bf 41.55\%\\
White, PGD 100 &37.39\% & 40.28\% & 38.98\%&\bf 40.78\%\\
White, PGD 200 &37.20\% & 40.24\% & 38.89\%&\bf 40.67\%\\
White, CW 30 &38.92\% &\bf 42.20\% & 40.75\% & 42.12\%\\
White, CW 100 &38.71\% & 41.88\% & 40.32\%&\bf 42.06\%\\\midrule
Transfer, PGD 5 & 68.03\% & 68.00\%  &\bf 69.78\%    & 69.00\%\\
Transfer, PGD 20 &62.77\% & 63.14\%  &\bf 64.70\%& 64.01\%\\
Transfer, PGD 100 &62.65\% & 63.12\%  &\bf 64.78\%& 64.15\%\\
Transfer, PGD 200 &62.73\% & 63.12\%  &\bf 64.69\%& 64.11\%\\
Transfer, CW 30 &63.89\% & 64.23\%  &\bf 65.65\% & 64.50\%\\
Transfer, CW 100 &63.77\% & 64.04\%  &\bf 65.43\% &64.35\% \\\bottomrule
\end{tabular}
\caption{Performance comparison of the adversarial training, adversarial training with feature regularization (AT-reg), adversarial training with attention (AT-att), and adversarial training with both (AT-att-reg) on CIFAR-10 using the 3-times wide ResNet network against PGD and CW adversaries.}
\label{table_cifar10_wide_results}
\end{center}
\end{table}

The evaluation results of aforementioned narrow and 3-times wide models on CIFAR10 are presented in Table~\ref{table_cifar10_narrow_results} and Tabel~\ref{table_cifar10_wide_results}. We find that roughly any $\lambda\in[0.01,1]$ works well for feature regularization. The reported results are obtained with $\lambda=1$ for AT-reg and $\lambda=0.1$ for AT-att-reg.
From the table, we see that all of the three proposed models have better adversarial robustness over the baseline model that only uses adversarial training, and both models with attention show improvement on the classification accuracy on natural examples as well. Note that the white box accuracy of the 3-times wide models are lower than the narrow models, which is due to overfitting of adversarial training~\cite{song2018improving}, and could be solved by early stopping.

We assess the effectiveness of feature regularization by comparing baseline to AT-reg in both Tables. The robustness is significantly improved especially under stronger attack with more steps, more than 5\% in some cases.

Next, by comparing the results of models with and without the attention module, we can see that attention contributes to both standard and adversarial accuracy such that the robustness improves by 1.5\% and at the same time standard accuracy by a similar margin. The attention structure not only favors robust features, it also relies heavily on features extracted from the spatial area that contains the actual object of concern. By suppressing the background clutter and misleading perturbations in irrelevant areas, the model with attention module more precisely learns the underlying distribution of the data which leads to better accuracy.

AT-att-reg takes advantage of both techniques and offers the most improvement in robustness and also better standard accuracy. We think attention and feature regularization complements each other. While regularization encourages the model to learn robust features from the input, attention assigns larger weights to robust features to make the model rely even more on them.

\subsection{Empirical Justification for Attention}
First of all, in order to show the advantage of our proposed non-linear attention model over the linear alignment model in Jetley et al.~\shortcite{jetley2018learn}, we train two narrow ResNet on natural CIFAR-10 samples with the two attention schemes respectively and the accuracy is 92.87\% for our attention and 91.34\% for the linear attention model.

\begin{figure}[t]
\begin{center}
\includegraphics[width=0.482\textwidth]{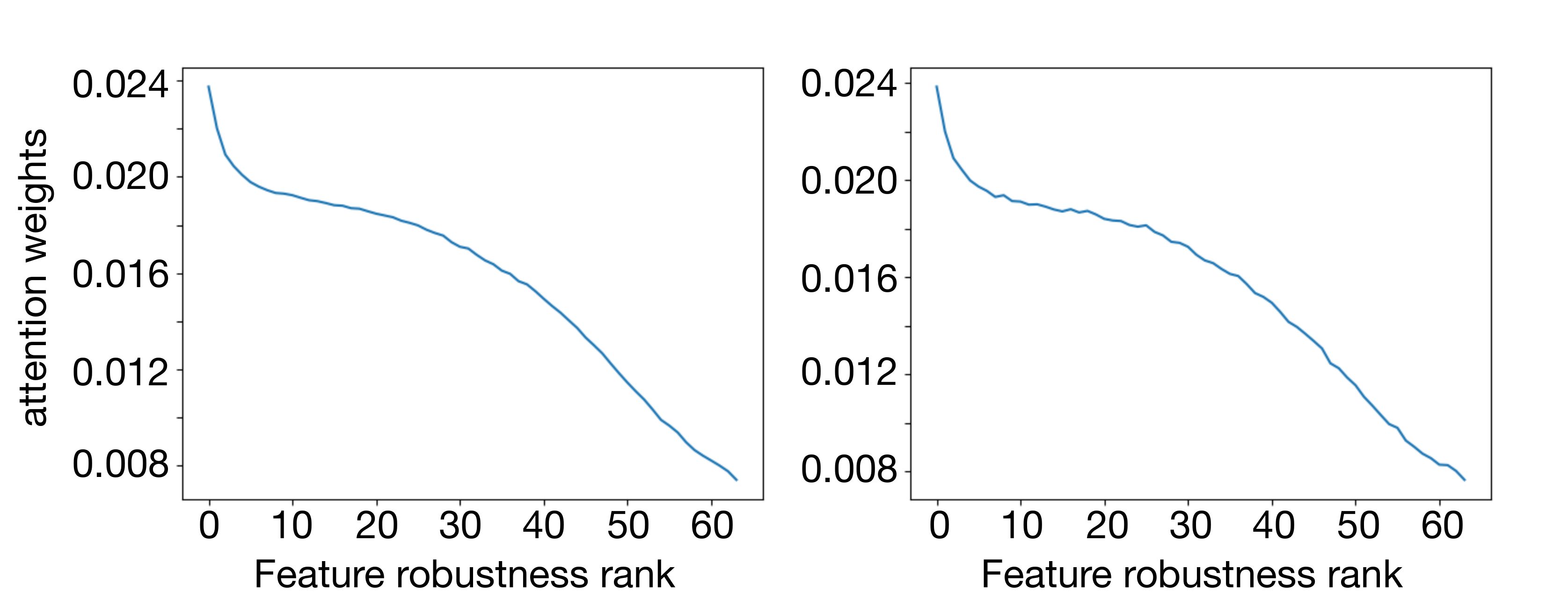}
\end{center}
\caption{The relationship between attention weights and feature robustness. The horizontal axis is the robustness rank, with 0 being the most robust and 63 the least robust, and the vertical axis is the corresponding attention weights. Left plot shows the results for training set and right plot is for test set of CIFAR-10.  \label{attention_weights}}
\end{figure}

Next, we empirically demonstrate that the attention module assigns larger weights to more robust features by examining the attention weights relative to the feature robustness. Figure~\ref{attention_weights} shows the relationship between the robustness of a feature and the magnitude of its assigned attention weight. The robustness measure we use is the $L_2$ distance between the learned features of a clean and an adversarial image, i.e. the smaller the distance between the features, the more invariant the feature is against input perturbations, therefore the feature is more robust. We rank the features accordingly and compute the average attention weights for each feature across all images. As shown in Figure~\ref{attention_weights}, our proposed attention mechanism indeed assigns larger weights to more robust features and less weights to non-robust features, so the model is more invariant to adversarial perturbation.

\begin{figure}[h]
\begin{center}
\includegraphics[width=0.48\textwidth]{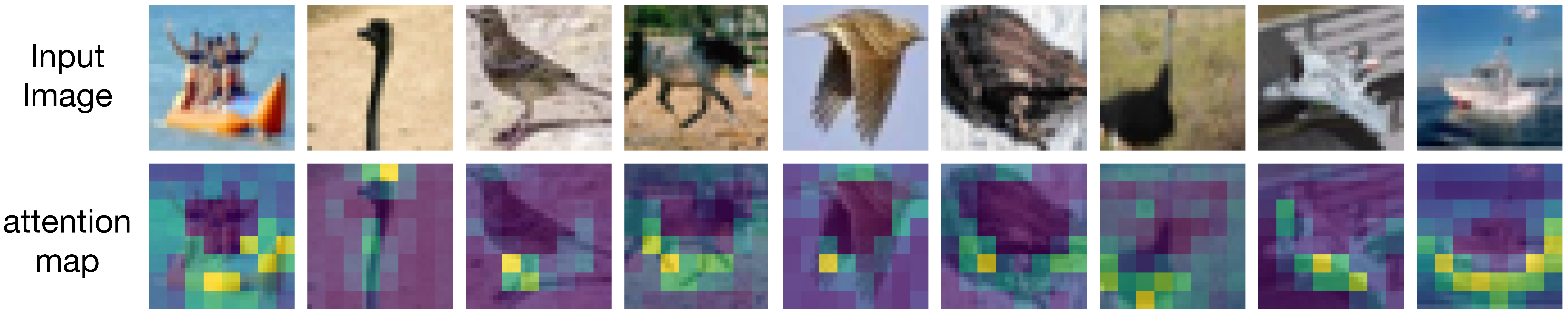}
\end{center}
\caption{The learned attention maps of our model. The first row are the input images and the second are the attention maps learned at residual block 4. \label{figure_attention_map}}
\end{figure}

Finally, we show the attention maps of our model in Figure~\ref{figure_attention_map} to visualize the attention weights. The attention maps focus sharply on the objects in the images, and the most relevant features like the head and legs of an animal and the wings of an airplane contribute more to the model's prediction and ignore the irrelevant background clutter.

\subsection{Gradient Map}
\label{subsubsection_gradient_map}
In this section, we study the gradient maps, which are the gradients of the cross-entropy loss with respect to input image pixels. The gradient maps directly indicate how the input features are utilized by a model for prediction and highlight the features which affect the loss most strongly. A large gradient on an input feature signifies a heavy dependence from the model. Human vision is robust against small input perturbations and the perception of an image reflects the reliance on input features. Therefore, models that depend on robust features will be better aligned with human vision, and the alignment can be used to evaluate the robustness of a model. Next, we show that the gradient maps generated from our model align better with the salient data characteristics.

\begin{figure}[t]
\begin{center}
\includegraphics[width=0.48\textwidth]{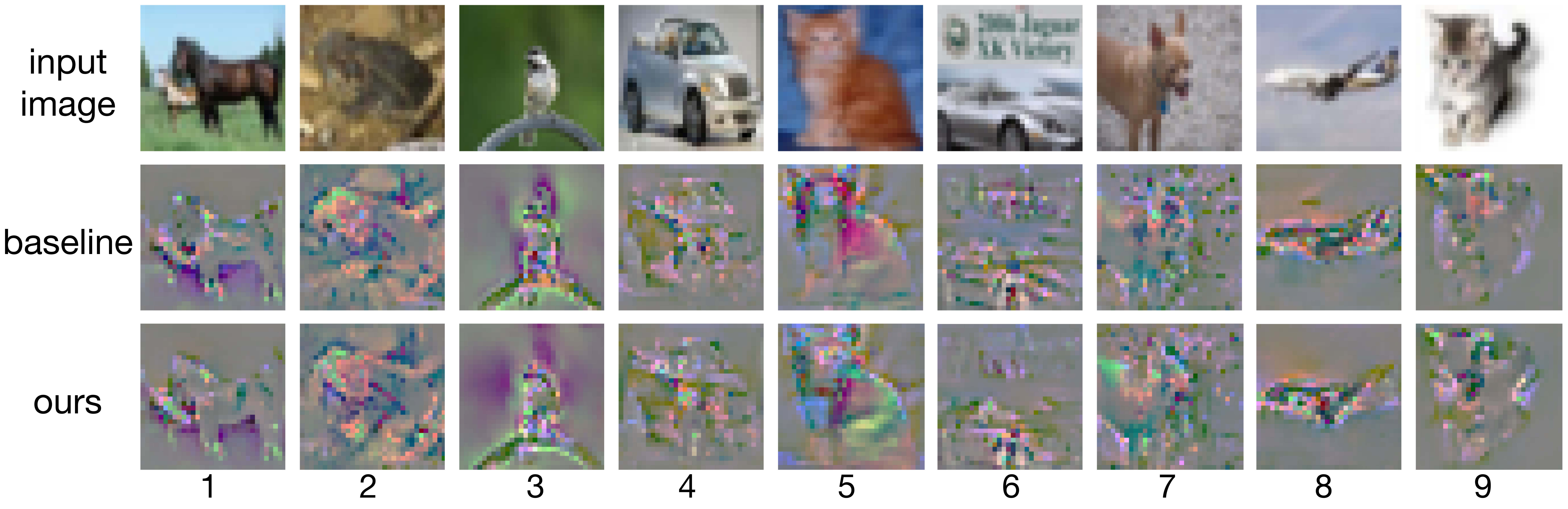}
\end{center}
\caption{Original CIFAR-10 images (top row) and corresponding gradient maps from Madry et al. [2017] model (mid row) and our model (bottom row). The raw gradients are clipped to within $\pm 3$ standard deviation and rescaled to lie in the [0, 1] range for visualization. \label{figure_grad_map}}
\end{figure}

First we present the qualitative result. Figure~\ref{figure_grad_map} shows the gradient maps from Madry et al.~\shortcite{madry2017towards} and our model. Overall, we note that gradient maps from both models are highly interpretable and align very well with the image features. Upon careful inspection, it is evident that the gradient maps generated from our model are better than the baseline. To point out a few, note that, in columns 2, 3, 6 and 8, our gradient maps have cleaner backgrounds and the gradients only focus on the objects; especially in column 6, the baseline model has large gradients on the text field in the background which is irrelevant to the class label (automobile), while in our model gradients in that area are much more suppressed. In columns 1, 5, and 9 the edges of the faces and heads of the animals are depicted clearer in our model. Human inspection which could be very subjective, therefore, we also introduce a quantitative evaluation method for gradient maps.

\begin{table}[h]
\begin{center}
\begin{tabular}{@{}l@{\hskip 0.01\textwidth}r@{\hskip 0.01\textwidth}r@{\hskip 0.02\textwidth}r@{\hskip 0.01\textwidth}r@{}}
\toprule
& \multicolumn{2}{l}{\bf With clipping} & \multicolumn{2}{c}{\bf Without clipping}\\
\cmidrule[0.05em](r{1.6em}){2-3} \cmidrule[0.05em](l{0em}){4-5}
\bf Method & \bf Train data & \bf Test data& \bf Train data&\bf Test data  \\ \midrule
Madry et al.  & 27.10\%  & 26.78\%& 28.60\% & 28.72\%    \\
Ours  & \bf 30.11\%  & \bf 30.32\%  & \bf 31.46\%  & \bf 31.59\%   \\
\bottomrule
\end{tabular}
\caption{Classification accuracy on the gradient maps from baseline and our model on both the training and test set of CIFAR-10. We run the experiment on gradient maps both with and without clipping to avoid the influence of gradient clipping.\label{table_grad_map_results}}
\end{center}
\end{table}

The problem we consider here is to decide how well the gradient maps align with the original images. The better they align, the more recognizable the gradient images are. A standard neural network extracts relevant features and make predictions based on them. When a gradient map is highly aligned with the original image, the neural net is able to identify more relevant features and thus the classification accuracy will be higher. Therefore, we can quantitatively compare the alignment by the classification accuracy of gradient maps. We use a pretrained network to classify the gradient maps for images in both the training set and the test set. 

We pretrained the same ResNet model as in Section~\ref{subsection_results_on_cifar10} with only natural training data of CIFAR-10. It achieves 88.79\% accuracy on the test set. The classification results are presented in Table~\ref{table_grad_map_results}. To avoid the possible influence of gradient clipping we also run the evaluation on raw gradients. As demonstrated by the classification accuracy, the gradient maps from our model express significantly better alignment with the original images. 

To summarize, both the qualitative and quantitative results show that the gradient maps from our model have better interpretability and alignment with the original images. It suggests that our model depends on robust features of the input images which explains the improved performance on both standard accuracy and adversarial robustness. 

\subsection{Results on CIFAR-100}
Here we present our results on the CIFAR-100 dataset. The experiment setup is the same as CIFAR-10 in Section~\ref{subsection_results_on_cifar10}. 
\begin{table}[h]
\begin{center}
\begin{tabular}{@{}l@{\hskip -0.02\textwidth}r@{\hskip 0.015\textwidth}r@{\hskip 0.015\textwidth}c@{\hskip 0.002\textwidth}r@{}}
\toprule
\bf Method & \bf Madry et al. & \bf AT-reg & \bf AT-att & \bf AT-att-reg  \\ \midrule
  Natural & 52.70\%    & 49.53\%    & \bf 53.67\%   & 50.66\%\\\midrule
White, PGD 5  & 25.14\% &26.99\% & 26.33\%  &\bf  27.76\%\\
White, PGD 20 &19.65\% & 23.16\%& 20.82\%& \bf 23.80\%\\
White, PGD 100 &19.47\% & 23.07\%& 20.59\%&\bf  23.62\%\\
White, PGD 200 &19.41\% & 22.96\%&20.53\% & \bf 23.62\%\\
White, CW 30 &18.64\% & 20.78\%& 19.39\%& \bf 20.88\%\\
White, CW 100 &18.61\% &20.63\% & 19.26\%& \bf 20.76\%\\\midrule
Transfer, PGD 5  & 35.37\%    & 35.10\% &\bf  35.95\%   & 35.17\%\\
Transfer, PGD 20 & 31.99\% & 31.88\%&\bf  32.48\%& 32.04\%\\
Transfer, PGD 100 & 32.03\% & 31.84\%& \bf 32.38\%& 32.00\%\\
Transfer, PGD 200 & 32.00\% & 31.80\%& \bf 32.37\%& 32.06\%\\
Transfer, CW 30 & 32.50\% & 31.96\%& \bf 32.75\%& 29.81\%\\
Transfer, CW 100 & 32.46\% & 31.86\%&\bf  32.74\% & 29.71\%\\\bottomrule
\end{tabular}
\caption{Performance comparison of the adversarial training, adversarial training with feature regularization (AT-reg), adversarial training with attention (AT-att), and adversarial training with both (AT-att-reg) on CIFAR-100 against PGD and CW adversaries.}
\label{table_cifar100_narrow_results}
\end{center}
\end{table}

From Table~\ref{table_cifar100_narrow_results}, similar with CIFAR-10, our model outperforms baseline by as much as 4\% on CIFAR-100 dataset.

\subsection{Comparison of Feature Regularization with Adversarial Logit Pairing}
Though feature regularization is more intuitive than adversarial logit pairing (ALP) as explained in Section~\ref{section_related_work}, their performance under PGD attack is similar. However, because ALP is based on logits, it is not as robust as feature regularization under CW attack~\cite{carlini2017towards}, which is based on logit margin.
\begin{table}[h]
\begin{center}
\begin{tabular}{@{}l@{\hskip 0.04\textwidth}r@{\hskip 0.015\textwidth}r@{\hskip 0.02\textwidth}r@{\hskip 0.015\textwidth}r@{}}
\toprule
\bf Method & \multicolumn{2}{c}{\bf ALP} & \multicolumn{2}{c}{\bf AT-reg}\\
\cmidrule[0.05em](r{1em}){2-3} \cmidrule[0.05em](l{0em}){4-5}
Adversary & CW 30 & CW 100& CW 30 & CW 100  \\ \midrule
Accuracy  & 39.17\%  & 39.15\%& \bf 42.96\% & \bf 42.87\%    \\
\bottomrule
\end{tabular}
\caption{Performance comparison of ALP and feature regularization on CIFAR-10 against CW adversaries. \label{table_alp}}
\end{center}
\end{table}

As shown in Table~\ref{table_alp}, AT-reg outperforms ALP by around 3.8\% against CW attacks.

\section{Conclusion}
In this paper, we propose feature prioritization and regularization to enhance both standard classification accuracy and adversarial robustness of a model over the baseline adversarial training approach. We provide empirical justifications for attention to show that it effectively favors robust features and focuses sharply on the region of interest. We then conduct quantitative and qualitative evaluation on gradient maps to show that they align perfectly with salient data characteristics, which further proves that our model heavily relies on the robust features.

\bibliographystyle{named}
\bibliography{ijcai19}

\begin{thebibliography}{}

\bibitem[\protect\citeauthoryear{Akhtar and Mian}{2018}]{akhtar2018threat}
Naveed Akhtar and Ajmal Mian.
\newblock Threat of adversarial attacks on deep learning in computer vision: A
  survey.
\newblock {\em arXiv preprint arXiv:1801.00553}, 2018.

\bibitem[\protect\citeauthoryear{Athalye and
  Carlini}{2018}]{athalye2018robustness}
Anish Athalye and Nicholas Carlini.
\newblock On the robustness of the cvpr 2018 white-box adversarial example
  defenses.
\newblock {\em arXiv preprint arXiv:1804.03286}, 2018.

\bibitem[\protect\citeauthoryear{Athalye and
  Sutskever}{2017}]{athalye2017synthesizing}
Anish Athalye and Ilya Sutskever.
\newblock Synthesizing robust adversarial examples.
\newblock {\em arXiv preprint arXiv:1707.07397}, 2017.

\bibitem[\protect\citeauthoryear{Athalye \bgroup \em et al.\egroup
  }{2018}]{athalye2018obfuscated}
Anish Athalye, Nicholas Carlini, and David Wagner.
\newblock Obfuscated gradients give a false sense of security: Circumventing
  defenses to adversarial examples.
\newblock {\em arXiv preprint arXiv:1802.00420}, 2018.

\bibitem[\protect\citeauthoryear{Carlini and Wagner}{2017}]{carlini2017towards}
Nicholas Carlini and David Wagner.
\newblock Towards evaluating the robustness of neural networks.
\newblock In {\em 2017 IEEE Symposium on Security and Privacy (SP)}, pages
  39--57. IEEE, 2017.

\bibitem[\protect\citeauthoryear{Dvijotham \bgroup \em et al.\egroup
  }{2018}]{dvijotham2018training}
Krishnamurthy Dvijotham, Sven Gowal, Robert Stanforth, Relja Arandjelovic,
  Brendan O'Donoghue, Jonathan Uesato, and Pushmeet Kohli.
\newblock Training verified learners with learned verifiers.
\newblock {\em arXiv preprint arXiv:1805.10265}, 2018.

\bibitem[\protect\citeauthoryear{Fawzi \bgroup \em et al.\egroup
  }{2018}]{fawzi2018analysis}
Alhussein Fawzi, Omar Fawzi, and Pascal Frossard.
\newblock Analysis of classifiers’ robustness to adversarial perturbations.
\newblock {\em Machine Learning}, 107(3):481--508, 2018.

\bibitem[\protect\citeauthoryear{Goodfellow \bgroup \em et al.\egroup
  }{2015}]{goodfellow2015explaining}
Ian~J. Goodfellow, Jonathon Shlens, and Christian Szegedy.
\newblock Explaining and harnessing adversarial examples.
\newblock {\em arXiv preprint arXiv:1412.6572}, 2015.

\bibitem[\protect\citeauthoryear{Goodfellow \bgroup \em et al.\egroup
  }{2016}]{goodfellow2016deep}
Ian Goodfellow, Yoshua Bengio, Aaron Courville, and Yoshua Bengio.
\newblock {\em Deep learning}, volume~1.
\newblock MIT Press, 2016.

\bibitem[\protect\citeauthoryear{Jetley \bgroup \em et al.\egroup
  }{2018}]{jetley2018learn}
Saumya Jetley, Nicholas~A Lord, Namhoon Lee, and Philip~HS Torr.
\newblock Learn to pay attention.
\newblock {\em arXiv preprint arXiv:1804.02391}, 2018.

\bibitem[\protect\citeauthoryear{Kannan \bgroup \em et al.\egroup
  }{2018}]{kannan2018adversarial}
Harini Kannan, Alexey Kurakin, and Ian Goodfellow.
\newblock Adversarial logit pairing.
\newblock {\em arXiv preprint arXiv:1803.06373}, 2018.

\bibitem[\protect\citeauthoryear{Kolter and Wong}{2017}]{kolter2017provable}
J~Zico Kolter and Eric Wong.
\newblock Provable defenses against adversarial examples via the convex outer
  adversarial polytope.
\newblock {\em arXiv preprint arXiv:1711.00851}, 2017.

\bibitem[\protect\citeauthoryear{Kurakin \bgroup \em et al.\egroup
  }{2016}]{kurakin2016adversarial}
Alexey Kurakin, Ian Goodfellow, and Samy Bengio.
\newblock Adversarial machine learning at scale.
\newblock {\em arXiv preprint arXiv:1611.01236}, 2016.

\bibitem[\protect\citeauthoryear{Liao \bgroup \em et al.\egroup
  }{2017}]{liao2018defense}
Fangzhou Liao, Ming Liang, Yinpeng Dong, Tianyu Pang, Xiaolin Hu, and Jun Zhu.
\newblock Defense against adversarial attacks using high-level representation
  guided denoiser.
\newblock {\em arXiv preprint arXiv:1712.02976}, 2017.

\bibitem[\protect\citeauthoryear{Madry \bgroup \em et al.\egroup
  }{2017}]{madry2017towards}
Aleksander Madry, Aleksandar Makelov, Ludwig Schmidt, Dimitris Tsipras, and
  Adrian Vladu.
\newblock Towards deep learning models resistant to adversarial attacks.
\newblock {\em arXiv preprint arXiv:1706.06083}, 2017.

\bibitem[\protect\citeauthoryear{Na \bgroup \em et al.\egroup
  }{2017}]{na2017cascade}
Taesik Na, Jong~Hwan Ko, and Saibal Mukhopadhyay.
\newblock Cascade adversarial machine learning regularized with a unified
  embedding.
\newblock {\em arXiv preprint arXiv:1708.02582}, 2017.

\bibitem[\protect\citeauthoryear{Prakash \bgroup \em et al.\egroup
  }{2018}]{prakash2018deflecting}
Aaditya Prakash, Nick Moran, Solomon Garber, Antonella DiLillo, and James
  Storer.
\newblock Deflecting adversarial attacks with pixel deflection.
\newblock In {\em Proceedings of the IEEE Conference on Computer Vision and
  Pattern Recognition}, pages 8571--8580, 2018.

\bibitem[\protect\citeauthoryear{Samangouei \bgroup \em et al.\egroup
  }{2018}]{samangouei2018defense}
Pouya Samangouei, Maya Kabkab, and Rama Chellappa.
\newblock Defense-gan: Protecting classifiers against adversarial attacks using
  generative models.
\newblock {\em arXiv preprint arXiv:1805.06605}, 2018.

\bibitem[\protect\citeauthoryear{Seo \bgroup \em et al.\egroup
  }{2016}]{seo2016progressive}
Paul~Hongsuck Seo, Zhe Lin, Scott Cohen, Xiaohui Shen, and Bohyung Han.
\newblock Progressive attention networks for visual attribute prediction.
\newblock {\em arXiv preprint arXiv:1606.02393}, 2016.

\bibitem[\protect\citeauthoryear{Simonyan and
  Zisserman}{2014}]{simonyan2014very}
Karen Simonyan and Andrew Zisserman.
\newblock Very deep convolutional networks for large-scale image recognition.
\newblock {\em arXiv preprint arXiv:1409.1556}, 2014.

\bibitem[\protect\citeauthoryear{Song \bgroup \em et al.\egroup
  }{2017}]{song2017pixeldefend}
Yang Song, Taesup Kim, Sebastian Nowozin, Stefano Ermon, and Nate Kushman.
\newblock Pixeldefend: Leveraging generative models to understand and defend
  against adversarial examples.
\newblock {\em arXiv preprint arXiv:1710.10766}, 2017.

\bibitem[\protect\citeauthoryear{Song \bgroup \em et al.\egroup
  }{2018}]{song2018improving}
Chuanbiao Song, Kun He, Liwei Wang, and John~E Hopcroft.
\newblock Improving the generalization of adversarial training with domain
  adaptation.
\newblock {\em arXiv preprint arXiv:1810.00740}, 2018.

\bibitem[\protect\citeauthoryear{Szegedy \bgroup \em et al.\egroup
  }{2013}]{szegedy2013intriguing}
Christian Szegedy, Wojciech Zaremba, Ilya Sutskever, Joan Bruna, Dumitru Erhan,
  Ian Goodfellow, and Rob Fergus.
\newblock Intriguing properties of neural networks.
\newblock {\em arXiv preprint arXiv:1312.6199}, 2013.

\bibitem[\protect\citeauthoryear{Tanay \bgroup \em et al.\egroup
  }{2018}]{tanay2018built}
Thomas Tanay, Jerone~TA Andrews, and Lewis~D Griffin.
\newblock Built-in vulnerabilities to imperceptible adversarial perturbations.
\newblock {\em arXiv preprint arXiv:1806.07409}, 2018.

\bibitem[\protect\citeauthoryear{Tram{\`e}r \bgroup \em et al.\egroup
  }{2017}]{tramer2017ensemble}
Florian Tram{\`e}r, Alexey Kurakin, Nicolas Papernot, Ian Goodfellow, Dan
  Boneh, and Patrick McDaniel.
\newblock Ensemble adversarial training: Attacks and defenses.
\newblock {\em arXiv preprint arXiv:1705.07204}, 2017.

\bibitem[\protect\citeauthoryear{Tsipras \bgroup \em et al.\egroup
  }{2018}]{tsipras2018there}
Dimitris Tsipras, Shibani Santurkar, Logan Engstrom, Alexander Turner, and
  Aleksander Madry.
\newblock There is no free lunch in adversarial robustness (but there are
  unexpected benefits).
\newblock {\em arXiv preprint arXiv:1805.12152}, 2018.

\bibitem[\protect\citeauthoryear{Wong and Kolter}{2018}]{wong2018provable}
Eric Wong and Zico Kolter.
\newblock Provable defenses against adversarial examples via the convex outer
  adversarial polytope.
\newblock In {\em International Conference on Machine Learning}, pages
  5283--5292, 2018.

\end{thebibliography}

\end{document}